\documentclass{article}

\usepackage{arxiv}

\usepackage[utf8]{inputenc} 
\usepackage[T1]{fontenc}    
\usepackage{amsmath}
\usepackage{hyperref}       
\usepackage{url}            
\usepackage{booktabs}       
\usepackage{amsfonts}       
\usepackage{nicefrac}       
\usepackage{microtype}      
\usepackage{cleveref}       
\usepackage{lipsum}         
\usepackage{graphicx}
\usepackage{natbib}
\usepackage{doi}
\usepackage{algorithm}
\usepackage{algorithmic}

\usepackage{multirow}

\title{Deep Explainable Learning \\ with Graph Based Data Assessing and Rule Reasoning}

\author{ Yuanlong Li \\
	Aliyun \\
	Alibaba Group\\
	Hangzhou, 310024, China \\
	\texttt{xunyuan.lyl@alibaba-inc.com} \\
	\And
	Gaopan Huang \\
	Aliyun \\
	Alibaba Group\\
	Hangzhou, 310024, China \\
	\texttt{gaopan.hgp@alibaba-inc.com} \\
	\And
	Min Zhou \\
	Xichang Steel and Vanadium Co., Ltd.\\
	Pangang Group \\
	Xichang, 615000, China \\
	\texttt{zhm9785@vip.sina.com} \\
	\And
	Chuan Fu \\
	Aliyun \\
	Alibaba Group\\
	Hangzhou, 310024, China \\
	\texttt{fuchuan.fc@alibaba-inc.com} \\
	\And
	Honglin Qiao \\
	Aliyun \\
	Alibaba Group\\
	Hangzhou, 310024, China \\
	\texttt{kenny.qhl@alibaba-inc.com} \\
	\And
	Yan He \\
	Aliyun \\
	Alibaba Group\\
	Hangzhou, 310024, China \\
	\texttt{yan.he@antgroup.com} \\
}

\renewcommand{\shorttitle}{Deep Explainable Learning}

\begin{document}
\maketitle

\begin{abstract}
Learning an explainable classifier often results in low accuracy model or ends up with a huge rule set, while learning a deep model is usually more capable of handling noisy data at scale, but with the cost of hard to explain the result and weak at generalization. To mitigate this gap, we propose an end-to-end deep explainable learning approach that combines the advantage of deep model in noise handling and expert rule-based interpretability. Specifically, we propose to learn a deep data assessing model which models the data as a graph to represent the correlations among different observations, whose output will be used to extract key data features. The key features are then fed into a rule network constructed following predefined noisy expert rules with trainable parameters. As these models are correlated, we propose an end-to-end training framework, utilizing the rule classification loss to optimize the rule learning model and data assessing model at the same time. As the rule-based computation is none-differentiable, we propose a gradient linking search module to carry the gradient information from the rule learning model to the data assessing model. The proposed method is tested in an industry production system, showing comparable prediction accuracy, much higher generalization stability and better interpretability when compared with a decent deep ensemble baseline, and shows much better fitting power than pure rule-based approach.
\end{abstract}


\section{Introduction}
For classification, with the powerful fitting capability of ensemble classifiers and deep learning, one can usually train a machine learning model with significantly low training loss in practice. However, these models often suffer from performance degeneration after deploying, even with a lot of approaches to avoid over-fitting applied. Can we utilize the great fitting power of deep learning ``at ease'' without worrying about learning something we should not?

Another problem along with powerful fitting technologies is low interpretability, which is often questioned when purely data-driven approaches are proposed to replace human expert knowledge-based decision procedures. In practice, tremendous elegant expert-knowledge based models were proposed to solve problems. With the development of powerful fitting technologies, a trend of replacing the expert-knowledge based models by purely data-driven based machine learning models is emerging, and shows great power to deal with large amount of noisy data, but with mere interpretability.

We face the above two problems in an industry data classification task, with human expert based explainable decision rules predefined. However, we cannot simply apply the rules to the data to get the classification result. The first reason is that the acquired data observations are noisy and if we directly apply the rules we end up with low accuracy results. The second reason is that the rules are also noisy, and sometime the human expert apply different rules (with different parameter settings) according to the data. To this end, we need an approach that can deal with noisy data and noisy expert rules.

Existing researches on Logic Programming(LP) \cite{evans2018learning} and explainable AI (XAI) \cite{bhatt2020explainable} are closely related to the above problem. LP tries to induct new rules from data; however, due to the large amount of noisy data, one can end up with very large rule set with limited classification accuracy, and the learned rules are different from the expert knowledge based rules. Existing approaches on XAI often attach explainable decorators to original models, or gradient based feature importance tracing to the deep models, yet these approaches cannot apply in our case to deal with the data noise and rule noise at the same time.

\begin{figure}[t!]
\centering
\includegraphics[width=0.6\columnwidth]{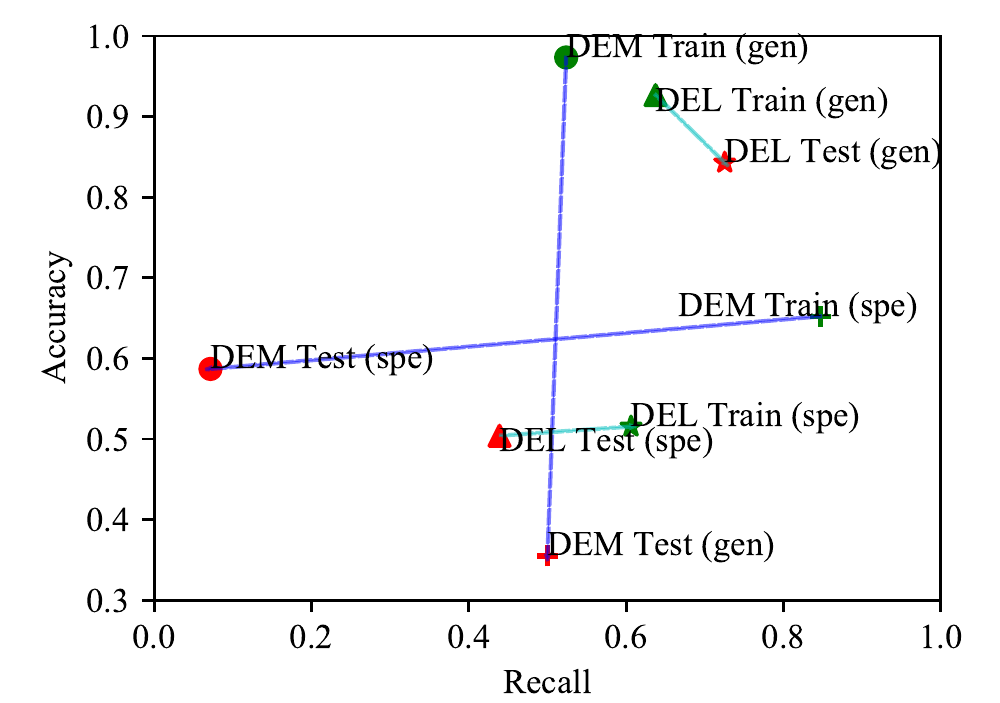} 
\caption{Performance changes when switching training and testing datasets. DEL is the proposed method and DEM is a deep ensemble baseline. ``gen'' and ``spe'' are the two datasets. Lines connect the performance scores and a shorter line indicates better stability.}
\label{intro_result}
\end{figure}

In this paper, we propose a deep explainable learning approach to tackle the above problem that combines the fitting power of deep learning and the interpretability of expert rules, with an end-to-end learning framework to handle the noise in both data and rules at the same time. Specifically, the approach simulates the decision process of a human expert, who evaluates the data, extracts certain key information from the large amount of data perceived, and then makes the decision according to some ``fuzzy'' rules according to the data. Following the above flow, we first propose a data assessing model to simulate the human expert ``data evaluation'' process, and by representing the data as a graph, a graph neural network based data assessing model is utilized to mine relations among data. Secondly, a rule learning model simulates the process that human expert adjusts rules according to the data received in practice, in which the noisy parameters in the rules are optimized. Finally, we propose a gradient linking search module to attach the rule learning model and the data assessing model into an end-to-end optimization flow, as the data assessing and rule learning process are correlated but the rule reasoning process can break the gradient back-propagation process. For example, a SQL query to compute a measurement of the data used in the rule formula can be non-differentiable. The proposed gradient linking search module can output target data assessing results ``recognized'' by the rules under optimization.

The proposed method is tested in an industry classification application, with noisy visual features as input and human expert crafted rules. We collect two datasets with different distributions, one dataset from daily production and the other one collected with stricter rules. Our experimental results show that the proposed method can achieve similar classification accuracy compared with a decent deep ensemble classifier when training and testing in a same dataset (split half to train and half to test), and shows much better stability when training and testing on different datasets, as shown in Figure \ref{intro_result}. And of course, our method has the advantage of interpretability as it can generate explainable results with the optimized rules.

\section{Related Works}
In this section, we concentrate on relevant studies of the proposed deep explainable learning approach, which includes Logic Programming(LP) and Explainable AI (XAI). More extensive surveys on LP and XAI can be found in works \cite{qu2020rnnlogic} \cite{bhatt2020explainable}.

\subsection{Logic Programming}

Adopted from research \cite{evans2018learning}, logic programming refers to a family of programming languages in which the central component is ``if-then'' rule. When LP is used in data mining, the generated rules can be easily interpreted by human being and potentially extended into data unseen in training. It is thus of great importance to study how to mine rule with LP, of which one research attracted a lot interests is inductive LP (ILP) \cite{gallaire1978logic} \cite{lavrac1994inductive}.

In a general form, ILP tries to solve the following problem.  Defined as a tuple $(P, N, B)$, where $P$ is a true fact set, $N$ is a false set, $B$ is a set of basic relations and known rules,  the ILP is to find a set of rules $R$, that produces the positive facts in $P$ and rejects the false facts in $N$.

ILP has been studied in a variety of forms, which differ in problem formulation, targeting domain, numerical representation and so on. Canonically, the ILP problem can be tackled by a path finding algorithm to solve the boolean satisfiability problem. We focus on recent researches that solve the problem in a more  ``differential'' manner. We categorize these researches into the following four classes: Markov random field approaches, reinforcement learning methods, end-to-end differentiable methods and logical neural networks.

\subsubsection{Markov Random Field Approaches.}
Approach Markov logic network (MLN) \cite{richardson2006markov} utilizes Markov random fields to represent formulae (rules), which is created first before inference and then a Gibbs sampling procedure is applied to calculate the probability of the target formula. The constants in the provided knowledge base, as inputs to the model, are modelled as boolean variables and all of these variables will be used in inference in the worst case. An alternative approach Hinge-Loss Markov random fields (HL-MRFs) \cite{bach2017hinge} can more efficiently model first-order logic without the need to model a full field for all constants (entities) in a knowledge base, with a template language probabilistic soft logic (PSL).

\subsubsection{Reinforcement Learning Methods.}
Reinforcement learning (RL) method is also proposed for ILP \cite{crouse2019deep}, which applies the standard actor-critic method to search for the desired relation combination to generate a target rule. As a rule can be formulated as a chain and the construction can only be rewarded when the chain is completed, RL method can face the sparse reward problem.

\subsubsection{End-to-End Differentiable Methods.}
Most of the recent advances fall into this category, like Neural Logic Machines \cite{dong2019neural}, RNNLogic \cite{qu2020rnnlogic}, CTP \cite{minervini2020learning}(based on NTP \cite{rocktaschel2017end}), DRUM \cite{sadeghian2019drum}, $\delta$ILP \cite{evans2018learning}, Neural LP \cite{yang2017differentiable} (based on Tensorlog \cite{cohen2016tensorlog}). All these approaches solve the rule generation problem in a neural learning manner to bypass the canonical path searching procedure.

Considering the method to represent the constants in the approach, these methods can be divided into two subcategories. The first kind is without an embedding method, like Neural LP,  $\delta$ILP and DRUM. Both DRUM and Neural LP follow the Tensorlog matrix style of relation representation, with rules formulated as a chain of matrices. For $\delta$ILP, a template is used to represent the candidate rules. 
The second kind is those with a specified word embedding method, such as NTP, CTP and RNNLogic. Entities are embedded with word embedding methods and relations are computed by the difference, like RotatE. In NTP/CTP, a tree template is used as user input and parametrized for optimization. In RNNLogic, an RNN is used to compute the score of a chain rule, in which the score of the relations and the path are calculated based on the relation/entity embedding vectors, while the rule path is generated in a VAE manner.

\subsubsection{Logical Neural Network.}
Logical neural network (LNN) \cite{riegel2020logical} is a framework that can accept a tree template of rule, and compute the probability of a candidate rule with numerically represented logic operators like conjunction, disjunction. An ILP approach based on LNN \cite{sen2021neuro} is proposed by maximizing the probability of the root node (corresponding to the target query or relation) parametrized by the leaf node settings. The LNN is different from other approaches in that it applies a graph neural network with nodes representing the logic operators, with 1-to-1 correspondence between neurons and the elements of logical formulae.

Existing LP methods are not applicable in our case, as we are given a predefined rule set, and the problem is to utilize the existing noisy rules in classification. Our rule learning method can be taken as a simplified version of the LNN method, which is enough for our rule learning part.

\subsection{Explainable AI}
XAI is extensively reviewed in \cite{bhatt2020explainable}, where existing approaches such as feature importance, counter-factual explanations, adversarial training are detailed. We note that compared to a rigorous science, XAI lacks of standard measurement metrics and a bunch of works \cite{2018Considerations} \cite{hase2020evaluating} \cite{doshi2017towards} \cite{miller2019explanation} discuss this problem and try to pose XAI into a rigorous science framework. Below we briefly review the related studies on XAI in a taxonomic hierarchy favouring more on their implementation styles. 

\subsubsection{Model Agnostic Approaches.}
Model agnostic approaches usually create a surrogate model, based on the original model but functioning as a standalone interpreting model. Typical methods like LIME \cite{ribeiro2016should}, SHAP \cite{lundberg2017unified} follow the additive feature attribution formulation and SHAP proposes a local surrogate model framework based on existing approaches. A detailed survey and comparison of local surrogate model based approaches can be found in \cite{lundberg2017unified}. For global surrogate model approaches, there are studies utilizing the original model to teach an explainable decision tree model \cite{frosst2017distilling}, and teaching a reduced-scale model that can output lower dimensional representations for interpretation \cite{2018Interpreting}.

\subsubsection{Model Related Approaches.}
Different to model agnostic approaches, we refer to those without building a surrogate model and those closely coupled with the original model as the model-related approaches. The first kind of these approaches is the gradient-based feature attribution methods, usually applied in computer vision tasks, to explain how a classier makes the decision with the input image. One typical method is Integrated Gradients \cite{sundararajan2017axiomatic}, in which the gradient against the input $x$ is integrated along the line from a reference point to the target $x$. A different approach \cite{kindermans2017learning} studies why these methods fail for simple linear model thus proposes a new method called PatternNet that can can work for certain simple cases. All those methods mentioned above try to visualize the key features in the original input image to prove that the classifier can recognize the key objects. The second kind of model-related approaches modifies the original neural network structure and adds special additional layers to make the model explainable. For example, xNN \cite{vaughan2018explainable} is a new architecture in which a linear layer is added after the input layer and an ensemble layer is added before the output layer, to generate interpretation by the embedding results of the first layer. On the other hand, NBDT \cite{wan2020nbdt} appends a decision tree layer to the end of general neural network; however, the proposed mechanism is based on class label hierarchy and cannot be generalized to other rules, which thus cannot be applied in our case. 

\subsubsection{Rule Based Approaches.}
Training a classification model with a series of rules falls obviously into XAI, which is similar to the above mentioned LP methods. Some researches focus purely on rule generation without LP are shown in \cite{cohen1995fast} \cite{letham2015interpretable} \cite{yang2017scalable}. In these works, rules are generated in a ``grow-and-prune'' manner, and often start from a pre-mined set of rules to reduce the model space.

There are a few approaches combining decision tree and rule mining, as shown in \cite{friedman2008predictive} \cite{wei2019generalized}. For these researches, rules are formulated into ensemble of rules like a decision tree, in which the rule ensemble needs to be optimized.

\section{Method}
\subsection{Problem Setup}
A set of classification rules is given based on human expert knowledge, for simplicity formulated into the following form. For a target class $C$, a set of classification rules $R$ is defined with multiple measurement formulae like $f_k(\boldsymbol{x}) < \theta_k$ for a sample $\boldsymbol{x}$, where $f_k(\boldsymbol{x})$ is a predefined measurement (can be a SQL query based measurement, with conditions defined by multiple formulae) and $\theta_k$ is the corresponding threshold for this measurement determined by human expert, $k=1,...,K$. A sample data $\boldsymbol{x}$ is classified as negative for class $C$ when $\boldsymbol{x}$ entails all formulae in $R$, otherwise positive when it fails at least one rule in $R$. Note that in practice the classification rules can be a rule tree, by converting it into the \textbf{conjunctive normal form}, we can greatly simplify the optimization process. Formulae without parameters to optimize are omitted here. The threshold vector $\boldsymbol{\theta}$ is the target parameter to be optimized.

Given noisy sample data $\boldsymbol{x} \in X$, where $\boldsymbol{x}=\{\boldsymbol{x}_{seq}, \boldsymbol{x}_{base} \}$  ($\boldsymbol{x}_{seq}$ is a sequence of observations and $\boldsymbol{x}_{base}$ is a vector of other features), the corresponding human-labelled classification result $y \in Y$, and the corresponding human-labelled critical feature $y_{feat} \in Y_{feat}$ (which are certain features of the key rows in $\boldsymbol{x}_{seq}$ when the human expert makes the classification decisions), the goal is to train a deep data assessing model $H$ that can filter noise in the observations,  and at the same time train a new rule set $R^\alpha$ with optimized parameter settings of $\boldsymbol{\theta}$ to output correct classification decisions.

\subsection{End-to-End Deep Explainable Approach}
\begin{figure*}[t]
\centering
\includegraphics[width=0.95\textwidth]{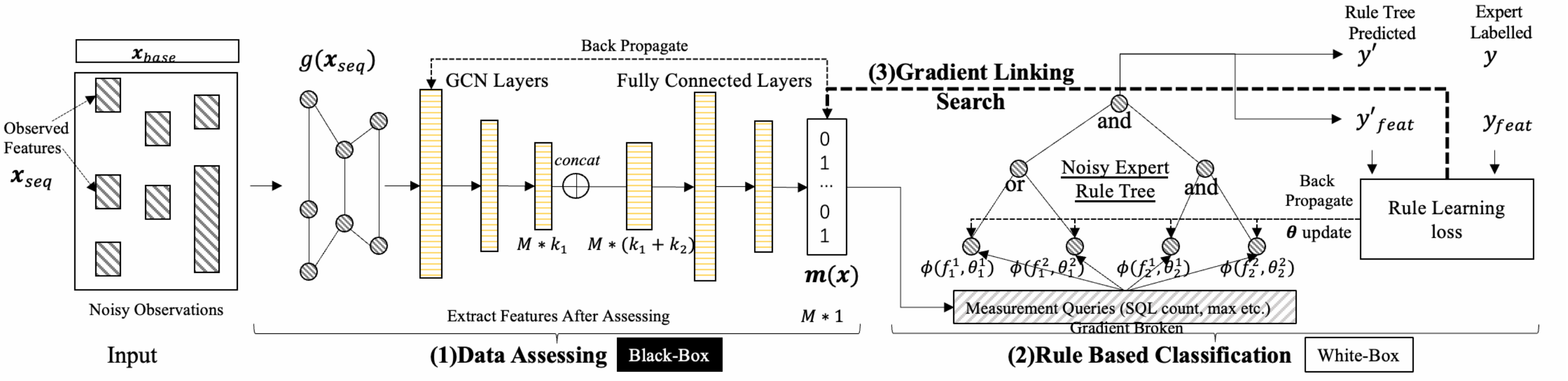} 
\caption{The proposed end-to-end deep explainable approach.}
\label{approach_fig}
\end{figure*}

We propose an end-to-end approach, that simulates the decision process with pre-defined rules of human expert facing with a large amount of noisy data. The approach can be divided into three parts: A data assessing model (which can simulate the process of extracting key features from the noisy observations), a rule learning model (which can simulate the process of the human expert adjusting rules according to the data), and a gradient linking search module, that can provide gradient information from the rule learning model back to the data assessing model, which is essential because the data query and computation process in rule based classification is not differentiable. As shown in Figure \ref{approach_fig}, the approach is a combination of black-box deep model and white-box rule-based classification model. The black-box deep model handles the large amount of noisy data, while the white-box rule model has great interpretability for practical application.

\subsubsection{Rule Learning Model.}

\begin{figure}[t]
\centering
\includegraphics[width=0.5\columnwidth]{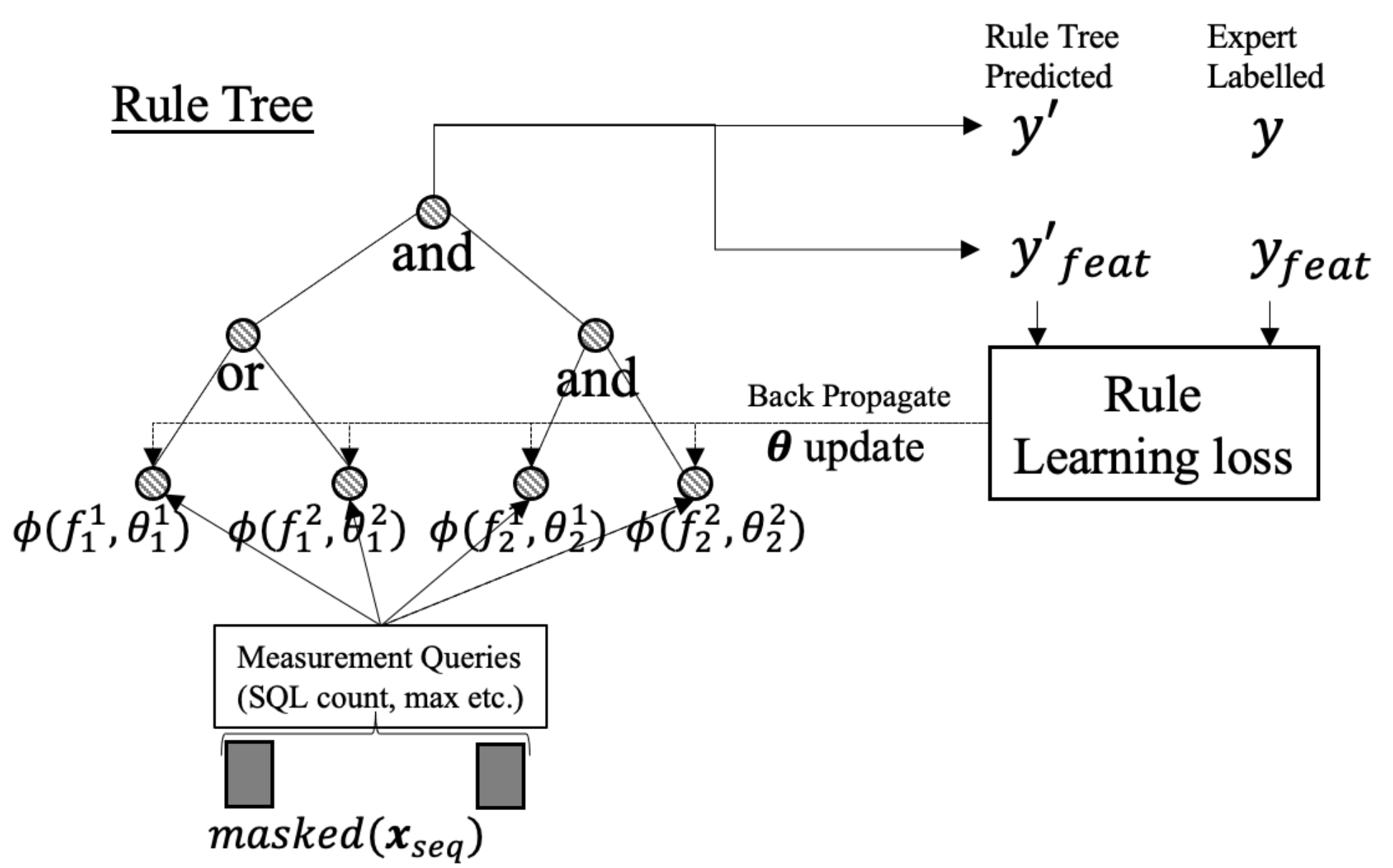} 
\caption{The rule learning procedure.}
\label{rule_learning_fig}
\end{figure}

We formulate the rule learning problem as a rule parameter optimization problem to induct a new threshold value for each measurement $f_k$, considering the distribution of the labelled complying samples and failed samples for the target class $C$. The target $\boldsymbol{\theta}$ should be able to classify the training samples the same as the human labels.

The inference for all thresholds for class $C$ can be formulated as following:

\begin{equation}
\max\limits_{\boldsymbol{\theta}} \sum_{\boldsymbol{x} \in X} [ \min\limits_{k=1,...,K}\phi(f_k(\boldsymbol{x}),\theta_k)]y_{C}(\boldsymbol{x}),
\label{eq:rule_goal}
\end{equation}
where 
\begin{equation}
\phi(f_k(\boldsymbol{x}),\theta_k) = -\textbf{sign}(f_k(\boldsymbol{x}) - \theta_k) ,
\end{equation}

and

\begin{equation}
y_{C}(\boldsymbol{x})=
\begin{cases}
1,& \text{$\boldsymbol{x}$ complying $C$}\\
-1,& \text{$\boldsymbol{x}$ failed $C$}
\end{cases}.
\end{equation}

Note that when $\boldsymbol{x}$ complies rules of class $C$, the objective tries to make sure that for every feature $f_k(\boldsymbol{x})$ achieves score 1; while $\boldsymbol{x}$ fails $C$, the objective tries to make sure that at least one rule is violated and a score -1 can be obtained.

To solve the above optimization problem, we build a rule tree based on the rules provided by human expert and then ``translate'' it into a neural network as shown in Figure \ref{rule_learning_fig}. The rule tree is constructed with leafs as the basic formulae, nodes as the logical computations (and/or). To ``translate'' the rule tree to a numerically tractable structure, we build neural network $G_{rule}$ that follows the rule tree structure with trainable parameter $\boldsymbol{\theta}$. We replace $\textbf{sign}$ to $tanh$ for smoother gradient, and utilize $\min$ and $\max$ operations to replace the logical ``and/or'' operations respectively. The model can be trained by the common batch-wise Adam optimizer with loss determined by the batch error shown in Equation \ref{eq:rule_goal}. We note that such simple approach is enough as we are not trying to optimize the rule tree structure or tuning logical operators.

We note that the parameter $\boldsymbol{\theta}$ may be easily over-fitted by the noise in $X$ and $Y$, and at the same time local optimal results can be introduced because of the exponentially growing search space for $\boldsymbol{\theta}$ with its length growing. In such case, the human-labelled critical features can be useful to guide the optimization. As when utilizing the classification rules, it is straightforward to output the key features (feature rows corresponding to the minimum score), we can compute the cross entropy loss of the predicted critical features and the ground-truth in training, which can then be used to regularizing the optimization process.  

Overall, the loss when optimizing the rule learning model is defined as following:
\begin{equation}
\begin{aligned}
L_{\boldsymbol{\theta}}(\boldsymbol{x}, y, y_{feat}) = & L_{focal}(G_{rule}(\boldsymbol{\theta}, \boldsymbol{x}),  y)+\\
& L_{ce}(y'_{feat}, y_{feat}),
\end{aligned}
\end{equation}
where $L_{focal}$ is the Focal loss \cite{lin2017focal} for the predicted labels,  $L_{ce}$ is the cross-entropy loss of the critical feature and $y'_{feat}$ is the predicted critical feature vector.

\subsubsection{Data Assessing Model.}
One common approach in logic programming is to learn a set of rules from a given dataset. When the dataset is at scale, one may face a problem that the generated rules may become too complicated to be explainable. That is different from what human being learn rules in the real world. To avoid overwhelming large rule set, human being often processes the noisy data first (assessing from multiple views), and then proposes the rules that should be utilized to certain key features. 
Such an ``assessing'' process is essential for explainable learning, and it affects the downstream classification task (the rules).  

\begin{figure}[t]
\centering
\includegraphics[width=0.6\columnwidth]{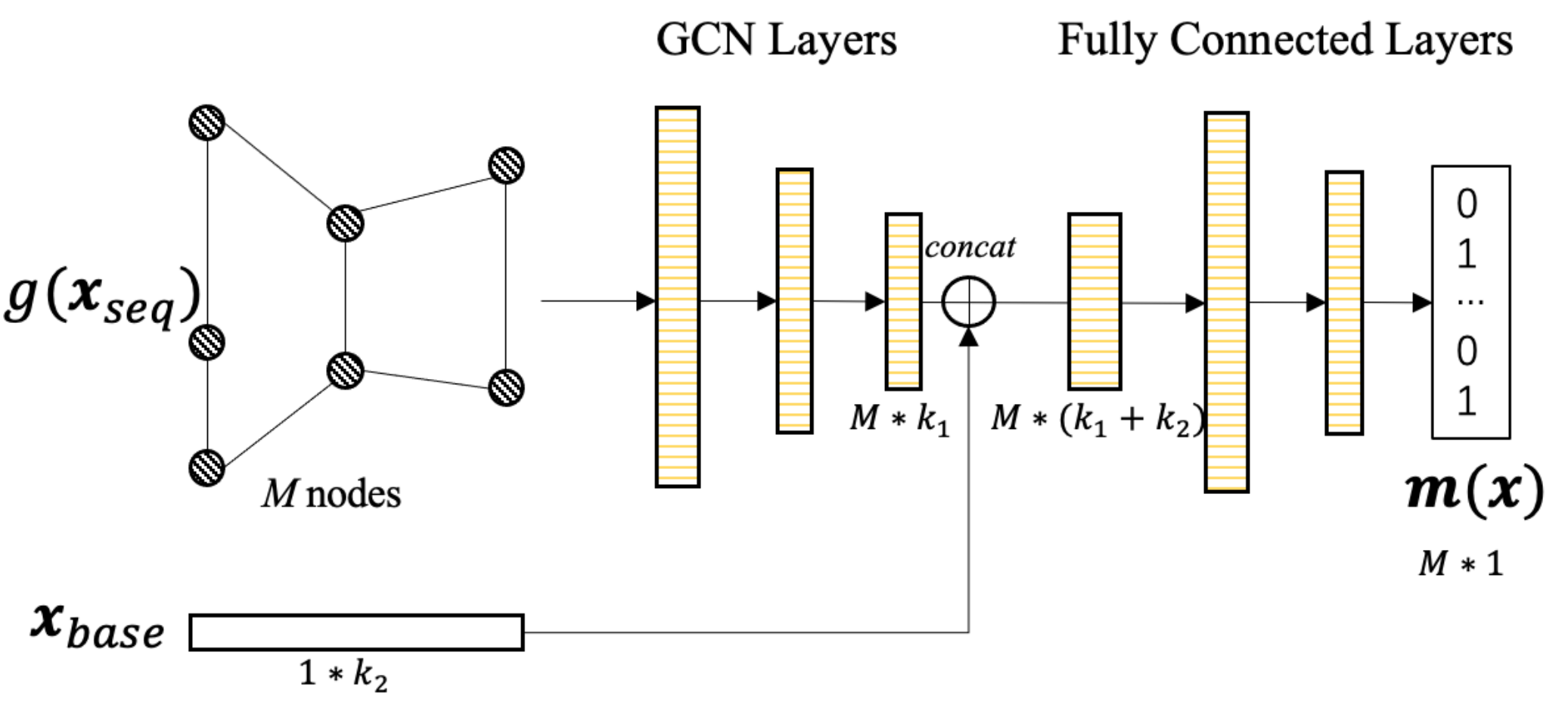} 
\caption{The data assessing model.}
\label{data_abs_fig}
\end{figure}

We propose a graph neural network (GNN) \cite{scarselli2008graph} based model $H$ to simulate the data ``assessing'' procedure as shown in Figure \ref{data_abs_fig}. Firstly, we map the feature sequence $\boldsymbol{x}_{seq}$ into a graph $g(\boldsymbol{x}_{seq})$ with each row of $\boldsymbol{x}$ as a node, and the edge of the graph is built according to certain relation between the feature rows. For example, in our test, each row of $\boldsymbol{x}_{seq}$ is an observation at certain position; edges are built when the distances of pairs of observations are lower than a threshold. The built graph is then processed by a GCN \cite{kipf2016semi} model, to output a sequence of embedding vectors $G_{assess}(g(x_{seq}))$, whose length is the same as $\boldsymbol{x}_{seq}$. The learned embedding is further concatenated with the base information $\boldsymbol{x}_{base}$, and then it is fed into a fully connected neural network to learn an assessing mask $\boldsymbol{m}(\boldsymbol{x})$, that will be used as a mask of $\boldsymbol{x}_{seq}$ before inputting to the rule network.

In summary, the data assessing model works as following:
\begin{equation}
H(\boldsymbol{x}_{seq}, \boldsymbol{x}_{base}) = F([G_{assess}(g(\boldsymbol{x}_{seq})), \boldsymbol{x}_{base}]),
\end{equation}
where $F$ is a fully connected neural network, $G_{assess}$ is a multiple-layer GCN network. The output $\boldsymbol{m}$ of model $H$ is a vector of the same length of $\boldsymbol{x}_{seq}$, and  $0 < m_i < 1$. When $m_i < 0.5$, the corresponding row of $\boldsymbol{x}_{seq}$ will be dropped in the downstream rule based classification.

The question left is how to train $H$.
If we directly connect the data assessing model with the rule learning model, the training of $H$ would require gradient back-propagation from the rule learning loss to the data assessing layers, which is not applicable in this case. The reason is that the gradient is broken because the rule tree has a lot of non-differentiable operations such as selection of certain rows of data to compute the measures in the rule. The method we fix such gradient broken problem will be detailed in the next subsection.

\subsubsection{Gradient Linking Search.}
We handle the above gradient broken problem by generating candidate masks $\boldsymbol{m}'$ to train the data assessing model with the loss defined as following:
\begin{equation}
L_{a}(\boldsymbol{x}, \boldsymbol{m}')= |H(g(\boldsymbol{x}_{seq}), \boldsymbol{x}_{base}) - \boldsymbol{m}'(\boldsymbol{x})|^2,
\end{equation}
where $\boldsymbol{m}'$ is the target mask to learn. In this manner, $H$ can be trained in a normal deep learning manner, the left problem is how to generate $\boldsymbol{m}'$ according to the rule learning result.

We propose a stochastic search procedure to generate $\boldsymbol{m}'(\boldsymbol{x})$. The search procedure starts from randomly generated paths $p$ in the feature graph $g(\boldsymbol{x}_{seq})$. The search will try to drop the first $s$ nodes in a path $p_i$ in step $s$, $s=1,...,S$ for all the candidate paths. The masked data will be processed by the rule tree model to generate the classification result, and if a correct prediction is made, the search will stop, otherwise the search will stop when all candidate steps on all paths are searched.

The final problem is how to determine the candidate paths. We generate paths by random walks in the feature graph, and amend it with the critical features in the rule tree prediction (at the current setting of rule parameter $\boldsymbol{\theta}$). In such manner, we utilize the potential relations among the features, and utilize the rule tree prediction feedback. 

\begin{figure}[t]
\centering
\includegraphics[width=0.7\columnwidth]{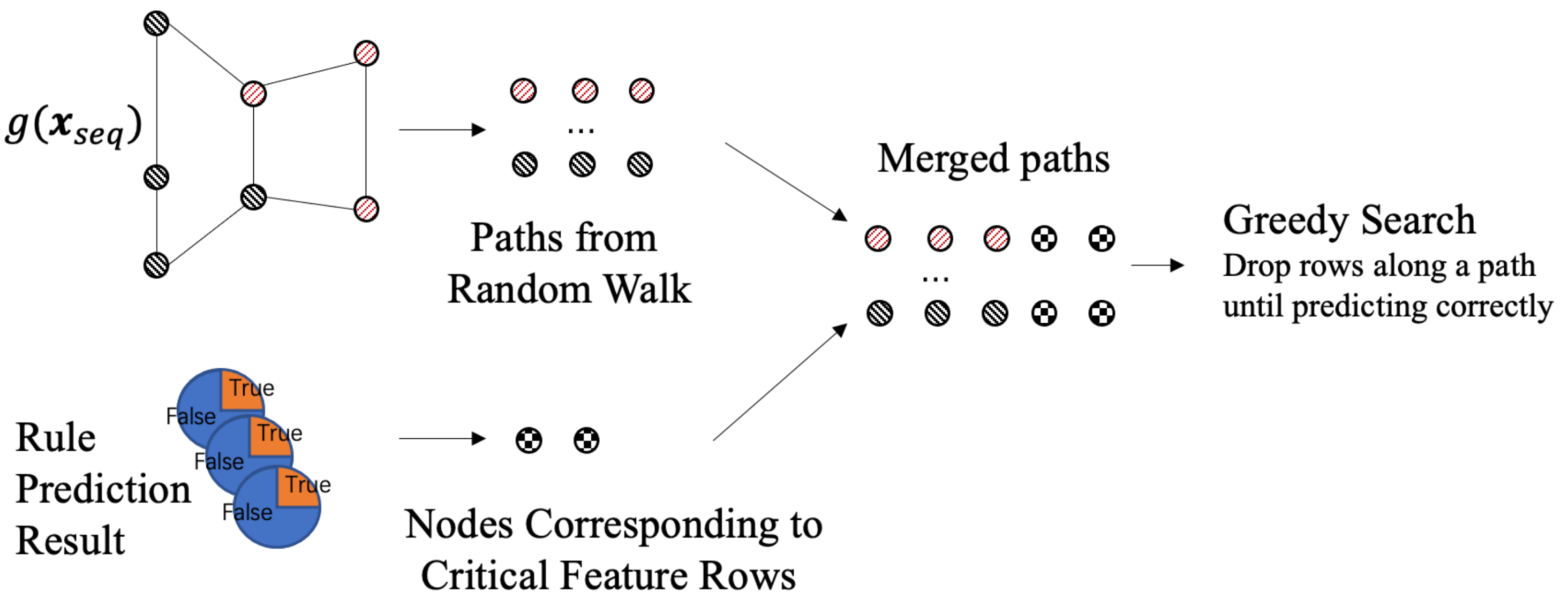} 
\caption{The gradient linking search procedure.}
\label{broken_handler_fig}
\end{figure}

To summary, the overall gradient linking search procedure is shown in Fig. \ref{broken_handler_fig}. For more details, please refer to the technical appendix.

\section{Experiments}

\subsection{Experimental Settings}
\subsubsection{Datasets and Models.}
The proposed approach is evaluated in an industry product quality evaluation system. In this system, each product is examined by a visual camera and a corresponding visual trace $\boldsymbol{x}_{seq}$ is generated. Length of $\boldsymbol{x}_{seq}$ can vary for different products, with length in range $[0, 20000)$. Also we can collect certain basic information $\boldsymbol{x}_{base}$ about the product. Human experts make product evaluation decisions (``qualified'' or ``unqualified'') according to the observations and a pre-defined rule set $r$,  and also output the key feature (rows of features from $\boldsymbol{x}_{seq}$) that is related to the decision.

We collected two datasets, from different human experts. The basic information of these two datasets are shown in Table \ref{tbl:dataset}. The first dataset $V_{general}$ is directly collected from daily production, with extremely unbalanced label distribution. The second dataset $V_{special}$ is a specially collected dataset that is much more balanced, but labelled by different experts and tends to be more strict on quality control (which can be considered as dataset generated with certain additional rules that are not included in the provided rule set). Their differences can be observed in the following experiments. 

\begin{table}[]
\centering
\begin{tabular}{ccc}
\hline
Dataset       & \begin{tabular}[c]{@{}c@{}}No. of \\ negative samples\end{tabular} & \begin{tabular}[c]{@{}c@{}}No. of \\ positive samples\end{tabular} \\ \hline
$V_{general}$ & 6584                                                               & 182                                                                \\ \hline
$V_{special}$ & 1372                                                               & 838                                                                \\ \hline
\end{tabular}
\caption{The two datasets used in experiments.}
\label{tbl:dataset}
\end{table}

\subsubsection{Baselines.}
In this paper, we compare our approach with two baselines: the basic rule (BR) based decision model and a decent deep ensemble model (DEM). BR utilizes the original expert-manipulated classification rules to classify the data samples. DEM is an ensemble method,  in which we first train a GCN based encoder to encode the sequential data into a fixed-length embedding vector, and then concatenate the embedding vector, the basic information of the product, and the decision results of the BR model into a final feature vector, which is then classified by a xgboost classifier. The parameter setting of xgboost classifier is tuned based on five-fold validations in the training dataset (more details can be found in the technical appendix). 

\subsubsection{Evaluation Metrics.}
All approaches are compared in the classify accuracy and the recall of the positive samples. As in our case the recall is more important than accuracy when the overall accuracy is sufficiently high, we train the model to optimize the recall score when the accuracy is higher than a pre-set accuracy threshold (92.5\% for $V_{general}$ and 50\% for $V_{special}$ as $V_{special}$ is much harder to classify). We denote such augmented recall as \textbf{Recall'}, which is set to the original recall score if the accuracy is higher than the target threshold, and set to negative recall score otherwise. In training we pick the model setup with the best \textbf{Recall'} which is used in testing.

\subsection{Results}
We compare the performance of the proposed approach with the baselines on the two datasets in different manners: ``\textbf{Closed test}'': train and test the algorithms on a same dataset (50\% as training data and the other 50\% as test data), ``\textbf{Open test}'': train on one dataset and test on the other dataset, for example, train on $V_{general}$, and test on $V_{special}$. Results are shown in Tables 2-5. 

We can see that the proposed DEL method outperforms the baseline DEM and BR in all the open tests, while achieves comparable results in the closed tests for $V_{general}$. DEM shows powerful fitting ability (100\% recall with very high accuracy, and performs much better for $V_{special}$ when rule set is incomplete) but with significant degeneration when migrate from training to test. BR performs poor and tends to be over strict, which is reasonable as the noise in $\boldsymbol{x}_{seq}$ can easily lead to false positive results.

An interesting observation is that the proposed DEL performs similarly when it is trained or tested on a dataset. For example the training recall on $V_{general}$ in Table \ref{tbl:result_open_gen} is similar to that of the test recall of $V_{general}$ in Table \ref{tbl:result_open_spe}. In comparison, DEM  shows significantly different results. A visualization of these gaps are shown in Figure \ref{intro_result}. This proves that DEL can be much more robust than general pure data-driven methods, which is essential for industry applications.

\begin{table}[]
\centering
\begin{tabular}{|c|c|cc|cc|}
\hline
Closed &  & \multicolumn{2}{c|}{Train(50\%)} & \multicolumn{2}{c|}{Test(50\%)} \\ \hline
 & \textbf{} & \multicolumn{1}{c|}{\textbf{Recall'}} & \textbf{Acc.} & \multicolumn{1}{c|}{\textbf{Recall}} & \textbf{Acc.} \\ \hline
\multirow{3}{*}{\begin{tabular}[c]{@{}c@{}}Train \\ $V_{general}$\end{tabular}} & DEM & \multicolumn{1}{c|}{1.0000} & 0.9997 & \multicolumn{1}{c|}{0.5238} & 0.9734 \\ \cline{2-6} 
 & BR & \multicolumn{1}{c|}{-} & - & \multicolumn{1}{c|}{0.9643} & 0.2814 \\ \cline{2-6} 
 & DEL & \multicolumn{1}{c|}{0.6429} & 0.9280 & \multicolumn{1}{c|}{\textbf{0.7143}} & \textbf{0.9137} \\ \hline
\end{tabular}
\caption{Closed test results on $V_{general}$.}
\label{tbl:result_close_gen}
\end{table}

\begin{table}[]
\centering
\begin{tabular}{|c|c|cc|cc|}
\hline
Closed &  & \multicolumn{2}{c|}{Train(50\%)} & \multicolumn{2}{c|}{Test(50\%)} \\ \hline
 & \textbf{} & \multicolumn{1}{c|}{\textbf{Recall'}} & \textbf{Acc.} & \multicolumn{1}{c|}{\textbf{Recall}} & \textbf{Acc.} \\ \hline
\multirow{3}{*}{\begin{tabular}[c]{@{}c@{}}Train \\ $V_{special}$\end{tabular}} & DEM & \multicolumn{1}{c|}{1.0000} & 0.8707 & \multicolumn{1}{c|}{\textbf{0.8469}} & \textbf{0.6522} \\ \cline{2-6} 
 & BR & \multicolumn{1}{c|}{-} & - & \multicolumn{1}{c|}{0.7309} & 0.4098 \\ \cline{2-6} 
 & DEL & \multicolumn{1}{c|}{0.5681} & 0.5022 & \multicolumn{1}{c|}{0.5556} & 0.5116 \\ \hline
\end{tabular}
\caption{Closed test results on $V_{special}$.}
\label{tbl:result_close_spe}
\end{table}

\begin{table}[]
\centering
\begin{tabular}{|c|c|cc|cc|}
\hline
Open &  & \multicolumn{2}{c|}{Train} & \multicolumn{2}{c|}{Test} \\ \hline
 &  & \multicolumn{1}{c|}{\textbf{Recall'}} & \textbf{Acc.} & \multicolumn{1}{c|}{\textbf{Recall}} & \textbf{Acc.} \\ \hline
\multirow{3}{*}{\begin{tabular}[c]{@{}c@{}}Train \\ $V_{general}$\end{tabular}} & DEM & \multicolumn{1}{c|}{0.5238} & 0.9734 & \multicolumn{1}{c|}{0.0716} & 0.5869 \\ \cline{2-6} 
 & BR & \multicolumn{1}{c|}{-} & - & \multicolumn{1}{c|}{0.7399} & 0.4122 \\ \cline{2-6} 
 & DEL & \multicolumn{1}{c|}{0.6374} & 0.9268 & \multicolumn{1}{c|}{\textbf{0.4391}} & \textbf{0.5041} \\ \hline
\end{tabular}
\caption{Open test results with $V_{general}$ as training data and $V_{special}$ as test data.}
\label{tbl:result_open_gen}
\end{table}

\begin{table}[]
\centering
\begin{tabular}{|c|c|cc|cc|}
\hline
Open &  & \multicolumn{2}{c|}{Train} & \multicolumn{2}{c|}{Test} \\ \hline
 &  & \multicolumn{1}{c|}{\textbf{Recall'}} & \textbf{Acc.} & \multicolumn{1}{c|}{\textbf{Recall}} & \textbf{Acc.} \\ \hline
\multirow{3}{*}{\begin{tabular}[c]{@{}c@{}}Train \\ $V_{special}$\end{tabular}} & DEM & \multicolumn{1}{c|}{0.8469} & 0.6522 & \multicolumn{1}{c|}{0.5000} & 0.3549 \\ \cline{2-6} 
 & BR & \multicolumn{1}{c|}{-} & - & \multicolumn{1}{c|}{0.9176} & 0.2742 \\ \cline{2-6} 
 & DEL & \multicolumn{1}{c|}{0.6062} & 0.5154 & \multicolumn{1}{c|}{\textbf{0.7253}} & \textbf{0.8429} \\ \hline
\end{tabular}
\caption{Open test results with $V_{special}$ as training data and $V_{general}$ as test data.}
\label{tbl:result_open_spe}
\end{table}

\subsection{Analysis}
\subsubsection{Learning Curve Example.}
\begin{figure}[]
\centering
\includegraphics[width=0.7\columnwidth]{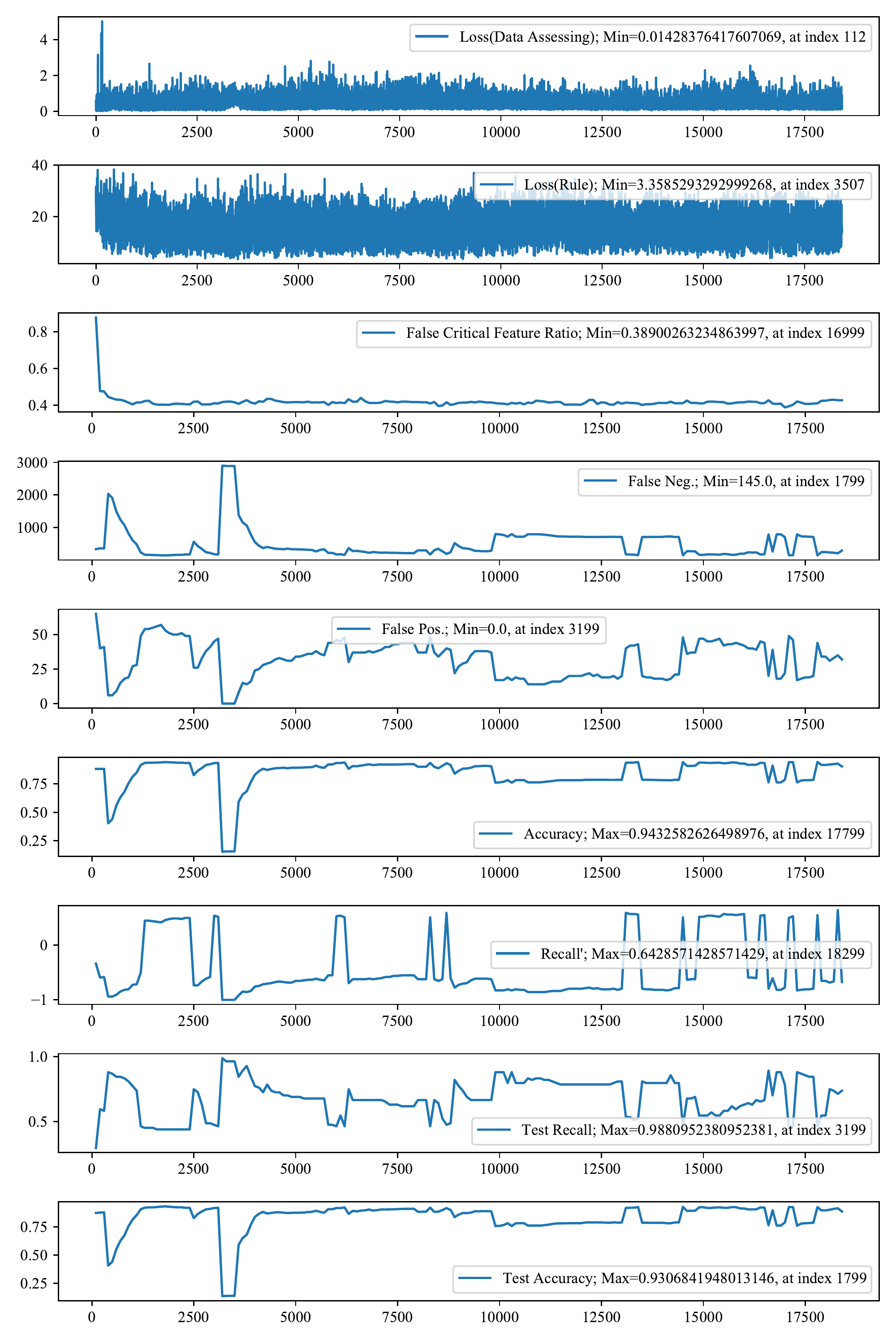} 
\caption{The learning curve when train and test with $V_{general}$. ``Loss(Data Assessing)'' is the training loss for the data assessing model. ``Loss(Rule) is the training loss for the rule optimizer. ``False Neg.'' denotes falsely predicted negative samples; ``False Pos.'' denotes falsely predicted positive samples. False critical feature ratio is the ratio that the critical features are not correctly predicted.}
\label{learning_curve}
\end{figure}

We present an example learning curve to demonstrate how DEL learns the optimized rule parameters and data assessing model in Figure \ref{learning_curve}. We observe that DEL tries to achieve a balance between ``False Neg.'' and ``False Pos.'' while the ``False Critical Feature Ratio'' is stably minimized. Note that the augmented recall will be smaller than 0 when the accuracy is below the target threshold. One can see that the test recall and accuracy curves basically follow the trend of the training curves, which we also observe in other experiments. 

\subsubsection{Ablation Study.}
We compare the standard DEL with two variants: one without the data assessing model, termed as NoDAM, and another without the critical feature loss, termed as NoCri.
Test results on NoDAM and NoCri compared with the standard DEL are shown in Tables 6-9. We observe that without the data assessing model, the performance degenerates by a moderate margin, which proves that the data assessing model can indeed help the rule fitting process. For NoCri, its performance is significantly degenerated after removing the critical feature loss, which is reasonable as the critical features can provide guiding information in the exponentially large search space. This proves that additional labelled information can be helpful for efficient rule optimization, which is a critical problem in ILP.

\begin{table}[]
\centering
\begin{tabular}{|c|c|cc|cc|}
\hline
Closed &  & \multicolumn{2}{c|}{Train(50\%)} & \multicolumn{2}{c|}{Test(50\%)} \\ \hline
 & \textbf{} & \multicolumn{1}{c|}{\textbf{Recall'}} & \textbf{Acc.} & \multicolumn{1}{c|}{\textbf{Recall}} & \textbf{Acc.} \\ \hline
\multirow{3}{*}{\begin{tabular}[c]{@{}c@{}}Train \\ $V_{general}$\end{tabular}} & NoDAM & \multicolumn{1}{c|}{0.5918} & 0.9386 & \multicolumn{1}{c|}{0.5595} & 0.9226 \\ \cline{2-6} 
 & NoCri & \multicolumn{1}{c|}{0.3061} & 0.9295 & \multicolumn{1}{c|}{0.2500} & 0.9256 \\ \cline{2-6} 
 & DEL & \multicolumn{1}{c|}{0.6429} & 0.9280 & \multicolumn{1}{c|}{\textbf{0.7143}} & \textbf{0.9137} \\ \hline
\end{tabular}
\caption{Ablation results for closed test with $V_{general}$.}
\label{tbl:result_close_gen_ab}
\end{table}

\begin{table}[]
\centering
\begin{tabular}{|c|c|cc|cc|}
\hline
Closed &  & \multicolumn{2}{c|}{Train(50\%)} & \multicolumn{2}{c|}{Test(50\%)} \\ \hline
 & \textbf{} & \multicolumn{1}{c|}{\textbf{Recall'}} & \textbf{Acc.} & \multicolumn{1}{c|}{\textbf{Recall}} & \textbf{Acc.} \\ \hline
\multirow{3}{*}{\begin{tabular}[c]{@{}c@{}}Train \\ $V_{special}$\end{tabular}} & NoDAM & \multicolumn{1}{c|}{0.4434} & 0.5013 & \multicolumn{1}{c|}{0.5160} & 0.5310 \\ \cline{2-6} 
 & NoCri & \multicolumn{1}{c|}{0.4942} & 0.5075 & \multicolumn{1}{c|}{0.5086} & 0.5060 \\ \cline{2-6} 
 & DEL & \multicolumn{1}{c|}{0.5681} & 0.5022 & \multicolumn{1}{c|}{\textbf{0.5556}} & 0.5116 \\ \hline
\end{tabular}
\caption{Ablation results for closed test with $V_{special}$.}
\label{tbl:result_close_spe_ab}
\end{table}

\begin{table}[]
\centering
\begin{tabular}{|c|c|cc|cc|}
\hline
Open &  & \multicolumn{2}{c|}{Train} & \multicolumn{2}{c|}{Test} \\ \hline
 &  & \multicolumn{1}{c|}{\textbf{Recall'}} & \textbf{Acc.} & \multicolumn{1}{c|}{\textbf{Recall}} & \textbf{Acc.} \\ \hline
\multirow{3}{*}{\begin{tabular}[c]{@{}c@{}}Train \\ $V_{general}$\end{tabular}} & NoDAM & \multicolumn{1}{c|}{0.6648} & 0.9257 & \multicolumn{1}{c|}{0.3294} & 0.4783 \\ \cline{2-6} 
 & NoCri & \multicolumn{1}{c|}{0.2747} & 0.9455 & \multicolumn{1}{c|}{0.1408} & 0.5118 \\ \cline{2-6} 
 & DEL & \multicolumn{1}{c|}{0.6374} & 0.9268 & \multicolumn{1}{c|}{\textbf{0.4391}} & 0.5041 \\ \hline
\end{tabular}
\caption{Ablation results for open test with $V_{general}$ as training data and $V_{special}$ as test data.}
\label{tbl:result_open_gen_ab}
\end{table}

\begin{table}[]
\centering
\begin{tabular}{|c|c|cc|cc|}
\hline
Open &  & \multicolumn{2}{c|}{Train} & \multicolumn{2}{c|}{Test} \\ \hline
 &  & \multicolumn{1}{c|}{\textbf{Recall'}} & \textbf{Acc.} & \multicolumn{1}{c|}{\textbf{Recall}} & \textbf{Acc.} \\ \hline
\multirow{3}{*}{\begin{tabular}[c]{@{}c@{}}Train \\ $V_{special}$\end{tabular}} & NoDAM & \multicolumn{1}{c|}{0.3735} & 0.5371 & \multicolumn{1}{c|}{0.5659} & 0.8625 \\ \cline{2-6} 
 & NoCri & \multicolumn{1}{c|}{0.4618} & 0.5086 & \multicolumn{1}{c|}{0.5824} & 0.5282 \\ \cline{2-6} 
 & DEL & \multicolumn{1}{c|}{0.6062} & 0.5154 & \multicolumn{1}{c|}{\textbf{0.7253}} & 0.8429 \\ \hline
\end{tabular}
\caption{Ablation results for open test with $V_{special}$ as training data and $V_{general}$ as test data.}
\label{tbl:result_open_spe_ab}
\end{table}


\section{Conclusions}
In this paper, we propose a deep explainable learning method, which combines the fitting power of deep model and the interpretability of expert rules. The proposed approach can tune expert rules and optimize the data-driven model in an end-to-end manner. The framework can give free rein to the powerful fitting capability of deep models, significantly enhance the model robustness and generalization capability, as shown in our experiments on different datasets in training and test. Our approach can shed light to tackling the robustness and interpretability challenges of deep models in real industry application.

\bibliographystyle{unsrtnat}
\bibliography{your_references.bib}

\section{Appendix}

\subsection{Pseudo Code of the DEL Framework}
The overall optimization framework of the proposed DEL is shown in Algorithm \ref{alg:overall_framework}. In the training procedure, training of the rule learning model has two stages. In the first stage, the data assessing model is not used in rule learning; while in the second stage, the data assessing model is incorporated. The reason to do so is that the learning of the data assessing model from random initialization requires a number of training epochs to output reasonable results. For the rule learning model, we also propose a global optimization module (a differential evolution optimizer from Scipy package) that do black-box random optimization aid for more stable optimization of $\boldsymbol{\theta}$. Parameter settings will be discussed in the next section.

One of the key component of DEL is the gradient linking search module (Algorithm \ref{alg:mask_generation}), in which the detailed greedy search procedure is shown in Algorithm \ref{alg:greedy_search}. With the gradient linking search module, we can handle basically any kind of rule classification model, with any-kind of non-differentiable operations. For example, the rule tree is not restricted to conjunctive normal form, although with which we can more easily trace the critical features. The search module itself can be implemented in parallel manner, which can be helpful in practice.

\begin{algorithm}[]
\caption{The DEL Optimization Framework}
\label{alg:overall_framework}
\textbf{Input}: Training dataset ($X^{train}$, $Y^{train}, Y_{feat}^{train}$), rule tree $R$.\\
\textbf{Parameter}: First stage epoch number $\Sigma_{1}=1000$, overall epoch number $\Sigma_{2}=40000$; rule learning rate $l_{rule}=0.0001$, data assessing model $H$  learning rate $l_{assess}=0.0001$; batch size $\beta=4$; global search frequency $\mu_{gopt}=50$, global optimization steps $\Sigma_{gopt}=5$; model validation frequency $\mu_{val}=100$.\\
\textbf{Output}: Validation results at each validation step, and the corresponding model snapshot.
\begin{algorithmic}[1] 
\STATE Let $t=0$.
\WHILE{$t < \Sigma_{2}$}
\STATE Generate train batch for rule learning with $\beta$ positive samples and $\beta$ negative samples.
\IF {$t < \Sigma_{1}$}
\STATE Train rule learning model with original $\boldsymbol{x}$ data with Adam optimizer (learning rate set to $l_{rule}$).
\ELSE
\STATE Train rule learning model with $\boldsymbol{x}$ masked by the mask generated by the data assessing model $H$ with Adam optimizer (learning rate set to $l_{rule}$).
\ENDIF
\IF {$t \% \mu_{gopt} = 0$}
\STATE Do global search for $\Sigma_{gopt}$ steps.
\ENDIF

\IF {$t \% \mu_{val} = 0$}
\STATE Do validation and output recall and accuracy scores for all samples in the \textbf{training} dataset.
\ENDIF

\STATE Generate train batch for data assessing model $H$ with $\beta$ positive samples and $\beta$ negative samples.

\STATE For these samples, generate candidate data assessing masks $\boldsymbol{m}'$.

\STATE Train data assessing model $H$ with Adam optimizer (learning rate set to $l_{assess}$).
\STATE $t \leftarrow t+1$.
\ENDWHILE
\end{algorithmic}
\end{algorithm}

\begin{algorithm}[]
\caption{Candidate Data Assessing Mask Generation Procedure}
\label{alg:mask_generation}
\textbf{Input}: Target data samples $X_{tar}$.\\
\textbf{Parameter}: Boolean flag $\eta$ (set to False in the first stage and True in the second stage).  \\
\textbf{Output}: Predicted masks $\boldsymbol{m}'$ for $X_{tar}$.

\begin{algorithmic}[1] 
\IF {$\eta$}
\STATE Generate masks with current data assessing model.
\STATE Update $X_{tar}$ with generated masks.
\ENDIF
\STATE Generate prediction results $y_{tar}'$ of $X_{tar}$ with the current rule network.

\FOR {$\boldsymbol{x} \in X_{tar}$}
\IF {$|y_{tar}'(\boldsymbol{x}) - y_{tar}(\boldsymbol{x})| < 1.0$}
	\IF {$\eta$}
	\STATE new mask $\boldsymbol{m}'(\boldsymbol{x})$ $\leftarrow$ The mask generated by the current $H$.
	\ELSE
	\STATE new mask $\boldsymbol{m}'(\boldsymbol{x})$  $\leftarrow$ All-one vector.
	\ENDIF
\ELSE
	\STATE new mask $\boldsymbol{m}'(\boldsymbol{x})$ $\leftarrow$ Greedy Search Procedure($\boldsymbol{x}$).
\ENDIF
\ENDFOR

\end{algorithmic}
\end{algorithm}

\begin{algorithm}[]
\caption{Greedy Search Procedure}
\label{alg:greedy_search}
\textbf{Input}: Training sample $\boldsymbol{x}$, current rule network $G_{rule}$.\\
\textbf{Parameter}: Greedy search steps $\chi=15$, number of tries in each step $\tau=10$.\\
\textbf{Output}: Target mask for sample $\boldsymbol{x}$.

\begin{algorithmic}[1] 
\STATE Classify sample $\boldsymbol{x}$ with the current rule network $G_{rule}$, return the predicted label $y'$ and critical features $y'_{feat}$.
\STATE $S$ $\leftarrow$ Randomly generate $\chi$ integers  smaller than $|\boldsymbol{x}_{seq}|$. 
\STATE Walk on graph $g(\boldsymbol{x}_{seq})$ to generate $\tau$ paths, each path of the length $\max(S)$. Amend $y'_{feat}$ to each path and randomly permute each path.
\FOR {$s \in S$}
	\FOR {$k=1,...,\tau$}
		\STATE $\boldsymbol{m}$ $\leftarrow$ All-one vector of the same length as $\boldsymbol{x}$.
		\STATE Set the first $s$ nodes in path $k$ of $\boldsymbol{m}$ into 0, and apply $\boldsymbol{m}$ to $\boldsymbol{x}$ (dropping the nodes with entries of $\boldsymbol{m}$ smaller than 0.5).
		\STATE Classify the modified sample $\boldsymbol{x}$ with the current rule network $G_{rule}$.
		\IF {Prediction is correct}
			\RETURN $\boldsymbol{m}$.
		\ENDIF
	\ENDFOR
\ENDFOR
\RETURN None
\end{algorithmic}
\end{algorithm}

\subsection{Results Production Setup}
More details to produce the experimental results are shown below. For the proposed DEL and its variants, we report the test results when the model achieves the best scores (augmented recall) on the training dataset for multiple tries (15 tries). Note that we do not need an additional validation dataset as the proposed algorithm is much more stable: the parameter settings in Algorithm \ref{alg:overall_framework}  are those generating the best training score. Note that parameter tuning for DEL can be easy, as we can simply try different parameter settings and then pick the setting that achieves the best score for the training set, without worrying about over-fitting. On contrary, for the baseline DEM we utilize five-fold cross-validations on the training dataset to select the best hyper-parameters for the xgboost classifier. We create 15 groups of hyper-parameter settings for the xgboost classifier and pick the setting that achieves best average score (augmented recall) in the five-fold cross-validations. 

Note that at the current stage we cannot open-source our code and the dataset as they are commercially confidential. We attach an example of the training and test log file and the script to extract the evaluation metrics in the code appendix.

\subsection{Neural Network Architecture Details}
In DEL, the rule learning network is constructed based on the rule tree. The inputs of the network (corresponding to the leaf nodes of the rule tree) are computed based on predefined rule formulae in form of queries on $\boldsymbol{x}_{seq}$. These queries are often ``counts'' or ``maximum values'' of certain kind of feature rows in $\boldsymbol{x}_{seq}$. For example, ``select count(*) from $\boldsymbol{x}_{seq}$ where feature type = $v$'', or ``select max(length) from $\boldsymbol{x}_{seq}$ where feature type = $v$''. For each leaf node, the query result and the rule parameter are used to compute the output of the leaf in the following normalized manner: 
\begin{equation}
o_{i}^{leaf} = \tanh((f_i(\boldsymbol{x}) - \theta_i)/Z_i),
\end{equation}
where $f_i(\boldsymbol{x})$ is the measurement of index $i$, $\boldsymbol{\theta}_i$ is the corresponding threshold parameter (trainable), and $Z_i$ is a constant normalization factor,  computed as the range of $f_i(\boldsymbol{x})$ on all training samples. With $o_{i}^{leaf}$, the following logic operations are then done in the following numerical manner: for ``and'' operation, we output the ``minimum'' value of the inputs; for ``or'' operations, we output the ``maximum'' value of the inputs.
Finally, the output of the rule network is an one-dimensional label. The key feature related to the prediction results is the feature rows incorporated in the queries corresponding to those leaf nodes with minimum value.

Details of the data assessing model are introduced below. The network $G_{assess}$ is a two-layered GCN, with embedding nodes 64 and 32 respectively. After GCN embedding, the output is then concatenated with $\boldsymbol{x}_{base}$ and fed into the fully connected network $F$, which is a two-layered neural network, with activations $\tanh$ and $sigmoid$ respectively. 

\subsection{Problems When Utilizing DEL to Other Applications }
In this section we discuss the problems one may face when utilizing the proposed DEL approach to other applications. 
The proposed DEL relies on two key conditions: the pre-defined expert rules and the labelled data. Expert rules are common to be seen in industry applications. These rules can be more complicated than the rules in our experiments, and rough rules are enough. The proposed rule learning model is capable of handling the rule noises as long as the rule learning problem can be formulated as certain parameter optimization problem. The proposed gradient linking module can deal with the none-differentiable operations that may be introduced by the rules. For the labels to train the model, we note that more labelled information generated when the human experts making the decision can be helpful, for example, the key features in our experiments. To summary, the proposed DEL approach can be easily adapted as a general framework to similar applications.


\end{document}